\PassOptionsToPackage{square,sort,comma,numbers}{natbib}
\documentclass{article}

\usepackage[preprint]{neurips_2026}

\usepackage[utf8]{inputenc} % allow utf-8 input
\usepackage[T1]{fontenc}    % use 8-bit T1 fonts
\usepackage{hyperref}       % hyperlinks
\usepackage{url}            % simple URL typesetting
\usepackage{booktabs}       % professional-quality tables
\usepackage{amsfonts}       % blackboard math symbols
\usepackage{nicefrac}       % compact symbols for 1/2, etc.
\usepackage{microtype}      % microtypography
\usepackage{xcolor}         % colors
\usepackage{microtype}

\usepackage{inconsolata}
\usepackage{algorithm}
\usepackage{algpseudocode}
\usepackage{booktabs} % For professional lines
\usepackage{makecell} % For line breaks inside cells
\usepackage{multirow} % For vertical centering
\usepackage[table]{xcolor} % Required for colored rows
\usepackage{times}
\usepackage{latexsym}
\usepackage{comment}

\usepackage{colortbl}
\usepackage{array}
\usepackage{caption}
\usepackage{listings}

\usepackage{graphicx}
\usepackage{amsmath}
\usepackage{tikz}     % Underlying graphics engine
\usepackage{pgfplots} % Required for the bar chart (\begin{axis})
\pgfplotsset{compat=1.18} % 

\newcommand{\reactagent}{ReAct}
\newcommand{\reactplanagent}{ReAct-Plan-Exec}
\newcommand{\spiralagent}{SPIRAL-Exec}

\definecolor{wingray}{gray}{0.88}
 
\newcommand{\best}[1]{\textbf{#1}}
\newcommand{\second}[1]{\underline{#1}}
\newcommand{\win}[1]{\cellcolor{wingray}#1}

\definecolor{bgcolor}{rgb}{0.97,0.97,0.97}
\definecolor{keywordcolor}{rgb}{0.0,0.2,0.6}
\definecolor{stringcolor}{rgb}{0.58,0.0,0.82}
\definecolor{commentcolor}{rgb}{0.25,0.5,0.35}
\definecolor{identifiercolor}{rgb}{0.1,0.1,0.1}

\lstdefinestyle{pythonstyle}{
    backgroundcolor=\color{bgcolor},
    commentstyle=\color{commentcolor}\ttfamily,
    keywordstyle=\color{keywordcolor}\bfseries,
    stringstyle=\color{stringcolor},
    identifierstyle=\color{identifiercolor},
    basicstyle=\ttfamily\small,
    breaklines=true,
    frame=single,
    rulecolor=\color{black!30},
    showstringspaces=false,
    tabsize=4,
    captionpos=b
}

\title{MCP-Cosmos: World Model-Augmented Agents for Complex Task Execution in MCP Environments}

\author{%
  Giridhar Ganapavarapu\\
  IBM, New York\\
  \texttt{giridhar.ganapavarapu@ibm.com} \\
  \And
  Dhaval Patel \\
  IBM, New York \\
  \texttt{pateldha@us.ibm.com} \\
}

\begin{document}
\maketitle
\begin{abstract}
The Model Context Protocol (MCP) has unified the interface between Large Language Models (LLMs) and external tools, yet a fundamental gap remains in how agents conceptualize the environments within which they operate. Current paradigms are bifurcated: \textit{Task-level planning} often ignores execution-time dynamics, while \textit{reactive execution} lacks long-horizon foresight. We present \textbf{MCP-Cosmos}, a framework that infuses generative World Models (WM) into the MCP ecosystem to enable predictive task automation. By unifying three disparate technologies, namely \textit{MCP}, \textit{World Model}, and \textit{Agent}, we demonstrate that a ``Bring Your Own World Model'' (BYOWM) strategy allows agents to simulate state transitions and refine plans in a latent space before execution. We conducted experiments using two strategies, namely ReAct and SPIRAL with 2 planning models and 3 representative world models over 20+ MCP-Bench tasks. We observed improvements in Agent's environment interaction KPI such as tool success rate and tool parameter accuracy. The framework also offers new metrics such as Execution Quality to generate new insights about the effectiveness of world models compared to baseline.
\end{abstract}

\section{Introduction}
The standardization of tool-augmented language models via the Model Context Protocol (MCP) has paved the way for sophisticated agentic workflows. However, existing evaluations of these agents reveal a fundamental tension between two architectural extremes. On one hand, planning-centric frameworks like \textit{TaskBench}~\cite{NEURIPS2024_085185ea} emphasize high-level decision-making based on static tool definitions but often fail to account for the stochasticity of real-world environments. On the other hand, execution-centric benchmarks such as \textit{MCP-Bench} rely on reactive paradigms, such as ReAct \cite{yao2023reactsynergizingreasoningacting}, in which the agent navigates the environment through interleaved observation and action. These reactive agents frequently suffer from ``horizon myopia,'' failing to anticipate downstream consequences, which leads to redundant tool calls or irreversible state failures. 

Simulating tool calls can help assessing impact of tool calls and mitigate possible failures in real-time action. This naturally requires application of world models into tool planning and evaluation methods to validate the use of these world models. In this paper, we propose \textbf{MCP-Cosmos}, a framework (as shown in Figure \ref{fig:architecture}) that integrates \textbf{World Models (WM)} into the MCP interaction loop. Our core hypothesis is that task-automation performance can be significantly enhanced by shifting from reactive execution to \textbf{Predictive Cognition}. By internalizing the environment's transition dynamics, represented as $P(s_{t+1} | s_t, a_t)$, where $a_t$ is an MCP tool call, agents can perform speculative look-ahead searches to refine their trajectories before committing to physical execution. Using this framework, we evaluate effectiveness of multiple world models in reactive and pro-active planning agents over 20+ MCP-Bench tasks and provide analysis. Further, we analyse gaps in current evaluation methodologies and propose new metric, Execution Quality to validate usefulness of world models, that is to avoid tool call failures and reduce redundant tool calls. Our primary contributions are as follows:

\begin{figure}[t!]
    \centering
    \includegraphics[scale=0.38]{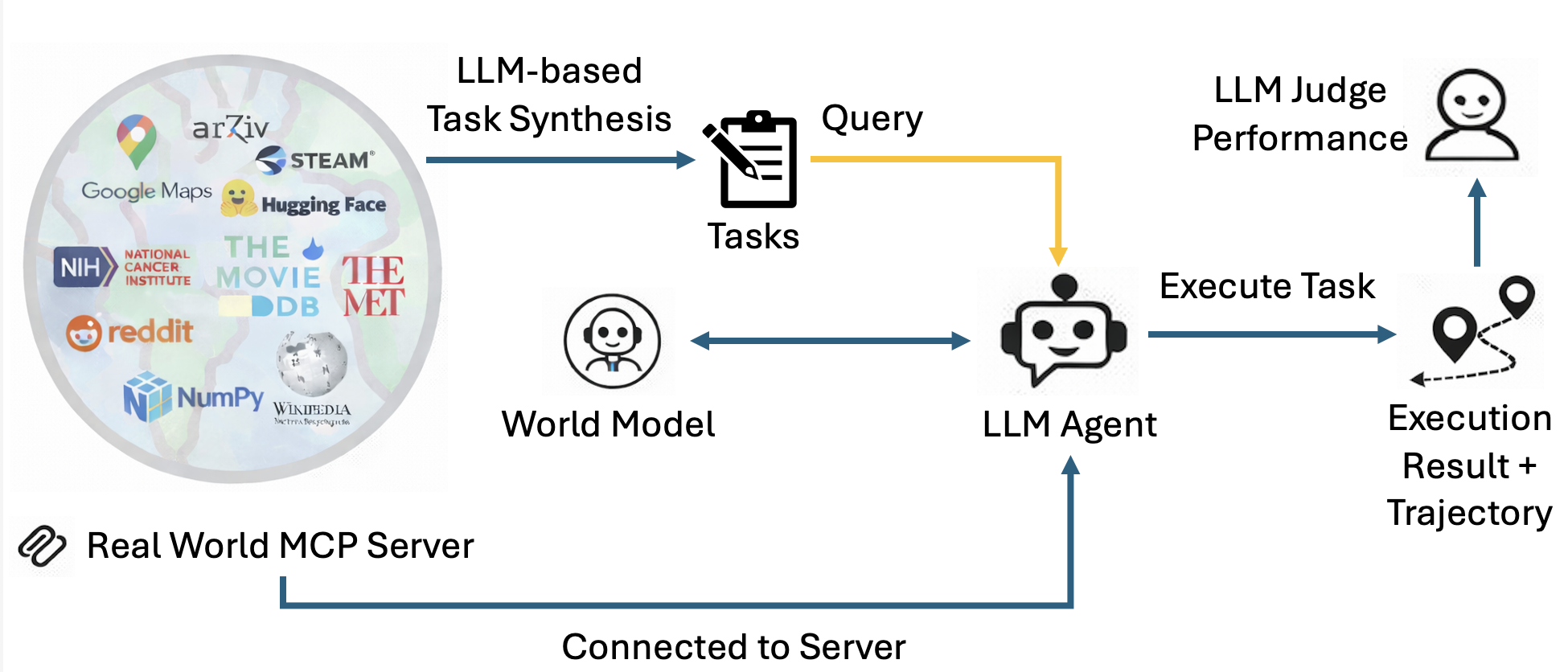}
    \caption{MCP-Cosmos : World Model-Augmented Agents for
Complex Task Execution in MCP Environments}
    \label{fig:architecture}
\end{figure}

\begin{itemize}
    \item \textbf{BYOWM Architecture:} We introduce a modular agentic strategy that enables the integration of heterogeneous World Models into the existing MCP ecosystem.
    \item \textbf{Benchmarking Predictive Efficiency:} We conduct a comparative analysis of world model infused agents against robust baselines over 300+ trajectories using existing task dataset, MCP tools and evaluation metrics in MCP-Bench \cite{wang2026mcpbench}.
    \item \textbf{Analysis on Gaps in Evaluation:} We provide detailed analysis on usefulness of current metrics and gaps in measuring effectiveness of world models.
    \item \textbf{New Evaluation Metrics:} We propose new evaluation metric - Execution Quality to measure ability of world model to guide efficient tool usage in agent planning. We claim that this new evaluation metric penalizes excessive use of tool calls than necessary for an input task. We emphirically prove the effectiveness of this metric in Section~\ref{sec:eval} and Section~\ref{sec:ablation}.
\end{itemize}

By treating MCP tool definitions as components of a predictive world model rather than simple API endpoints, we demonstrate that agents can achieve higher success rates in tool calling. This shift enables a significant portion of the reasoning process to occur in the latent ``simulated'' world, thereby preserving execution resources and improving overall system robustness.

\begin{figure}[t!]
    \centering
    \includegraphics[width=\columnwidth]{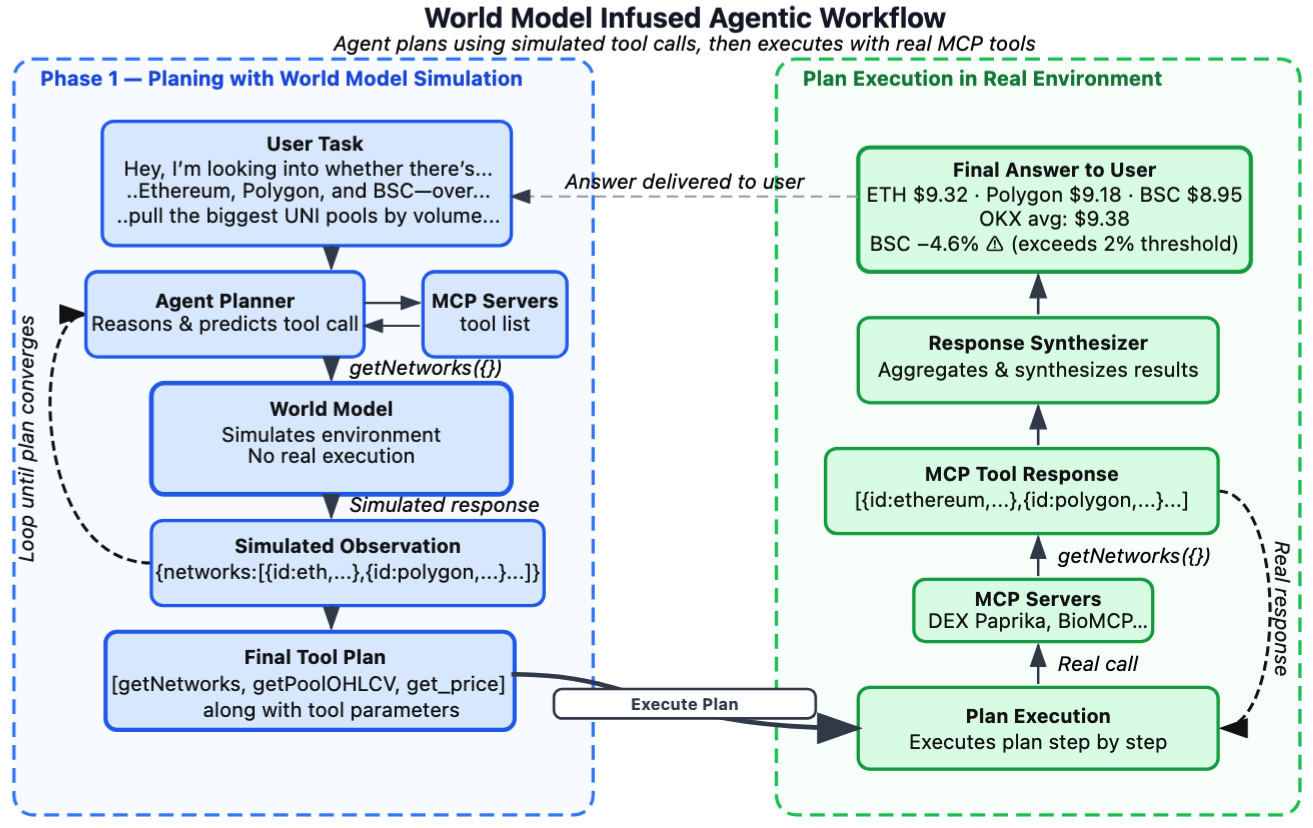}
    \caption{Sample workflow demonstrating simulation based agentic planning, execution and final answer synthesis.}
    \label{fig:wmflow}
\end{figure}

\section{World Model Infused Planning and Execution in MCP}

Agents orchestrate MCP tools using simulation and answer the user request in two phases as demonstrated in Figure \ref{fig:wmflow}. In phase-1, agents leverage world model simulations and plan complete tool orchestration proactively without interacting with actual MCP tools and environments. In phase-2, agents carry out the selected plan in the actual environment. We formalize the whole approach of integrating world models into multi-turn planning and execution in Algorithm~\ref{appendix:algorithm}.

\subsection{World Model Simulations}
We provide abstract API definition for tool planning world models in Listing~\ref{lst:wmabsapi}. Each world model can be initialized using the model name and additional parameters specific to that model. For a given tool call string and user request as input, the \texttt{simulate} method simulates the environment in latent space and returns the simulated tool response as observation. World models can use this method to implement various ways of creating simulation environments. Agent World Model introduced fully synthetic environment generation pipeline for MCP use cases in \cite{wang2026agent} and produced AWM 4B, 8B and 14B world models. We developed world model class for AWM 4B model implementing the WorldModel abstract definition.

\begin{lstlisting}[style=pythonstyle, language=Python, caption={Abstract WorldModel Class}, label={lst:wmabsapi}]
class WorldModel(ABC):
        
    def __init__(
        self,
        model_name: str,
        **kwargs
    ):
        """Initialize world model."""
        pass
    
    async def simulate(
        self,
        tool_call: str,
        user_request: str,
        context: Optional[str] = None
    ) -> Dict[str, Any]:
        """Simulate execution of a tool call."""
        pass
    
    def __repr__(self) -> str:
        """Return string representation of world model."""
        pass
    
    def to_dict(self) -> Dict[str, Any]:
        """Convert world model to dictionary representation."""
        pass
\end{lstlisting}

\subsection{Planning with World Model Simulation}

As shown in Figure~\ref{fig:wmflow}, agent planner utilizes world model to simulate potential action sequences without incurring the cost of real-world execution. The agent iteratively generates tool calls and revises the plan using the simulated observation until reaches a viable plan or specified termination criteria. An example user task, simulated observation for getNetworks tool call and final tool plan are provided in Figure~\ref{fig:wmflow}. These action and simulated observation pairs are accumulated in the world model trajectory, allowing the agent to explore multiple potential paths efficiently. The abstract API definition for infusing world models into agents is provided in Listing~\ref{lst:wminfusedagentapi}. The \texttt{WMInfusedAgent} takes initialized \texttt{world\_model} instance as input and uses for simulation in \texttt{execute} method.

\begin{lstlisting}[style=pythonstyle, language=Python, caption={WMInfusedAgent Class}, label={lst:wminfusedagentapi}]
class WMInfusedAgent:
    def __init__(
        self,
        generation_model_name: str,
        max_iterations: int,
        execute_w_revision: bool,
        world_model: Optional[Any] = None,
    ) -> None:
        pass
    
    async def execute(self, task: str) -> Dict[str, Any]:
        "return {"solution": solution, ...}
\end{lstlisting}

Finally, we select optimal plan from the world model trajectory using using non-deterministic policy models such as LLM or any deterministic selection algorithm such as reward based MCTS \cite{zhang2025spiralsymbolicllmplanning}. The selected final plan represents sequence of tool calls to be executed along with their parameters.

\subsection{Plan Execution in Real Environment}

The execution phase carries out execution of selected plan in the actual environment. The agent executes real MCP tools and receives the real observations. If an action fails during execution, the algorithm can optionally invoke a plan adjustment mechanism to modify the remaining plan, demonstrating robustness to execution failures. Due to the expensive nature of these revisions, we excluded the plan revision step from our benchmarking pipeline. At each step, successfully executed action-observation pairs are recorded in the execution trajectory $\tau$.

Upon completing the execution of plan, we generate final answer using summarization or synthesize techniques. The algorithm returns three key outputs: the final answer, the execution plan $\mathcal{P}$, and the complete execution trajectory $\tau$, providing full transparency into both the planning and execution processes. We formalized algorithm in Appendix~\ref{appendix:algorithm}.

\section{Experimental Setup and Results}

In establishing the evaluation foundation for \textbf{MCP-Cosmos}, we evaluate two primary alternatives for task grounding: \textbf{MCP-Universe} \cite{luo2025mcpuniversebenchmarkinglargelanguage} and \textbf{MCP-Bench} \cite{wang2026mcpbench}. While MCP-Universe serves as a broad repository of protocol-compliant resources, we use \textbf{MCP-Bench} as our primary dataset because of its innovative, ecosystem-scale design and focus on complex, real-world utility. More details on dataset are provided in \ref{datasetselection}. We extended MCP-Bench benchmark framework with three agentic architectures and three World-Models. More details on agents and models are provided in \ref{agentselection} and \ref{modelselection} respectively.

\subsection{Dataset and Scenario Selection}
\label{datasetselection}

The selection of \textbf{MCP-Bench} is driven by several methodological advantages inherent to its rigorous evaluation framework:
\begin{itemize}
    \item \textbf{Ecosystem-Scale Complexity:} It evaluates agents across 28 live MCP servers and 257 cross-domain tools.
    \item \textbf{Realistic Grounding:} It utilizes fuzzy-instruction tasks that challenge LLM grounding across multi-step processes.
    \item \textbf{Rigorous Evaluation:} The framework integrates rule-based and judge-based metrics with high human agreement.
    \item \textbf{Difficulty Stratification:} Tasks are categorized by structural complexity, using server counts as proxies for difficulty.
    \item \textbf{Diagnostic Depth:} It provides a taxonomy of planning failure modes, enabling granular performance analysis.
\end{itemize}

For the \textbf{MCP-Cosmos} evaluation suite, we specifically selected 2 and 3-server scenarios that emphasize cross-domain tool dependencies. These serve as our primary difficulty proxy, testing the agent's ability to maintain state across disparate tool domains and evaluate stability in handling fuzzy instructions. These multi-server scenarios evaluate World-Model effectiveness in predicting multiple tool outputs. Given that tool calls to some MCP servers incur costs after free tier limits, we curated a \textbf{cost-effective} subset of 24 tasks. A full list of specific tasks and detailed server mappings is provided in Appendix~\ref{sec:appendix_tasks}. The resulting experiments cover 300+ trajectories with 12 unique task types.

\subsection{Agent Architectures}
\label{agentselection}
We designed a comprehensive benchmark evaluation framework comparing three different agent architectures on multi-server MCP (Model Context Protocol) tasks:

\begin{enumerate}
    \item \textbf{\reactagent (Baseline)}: Standard reasoning and acting agent without World-Model capabilities, serving as the strongest baseline for comparison.
    
    \item \textbf{\reactplanagent}: Extended ReAct agent to generate pro-active plan using action generation and world model simulation, and execute the final plan.
    
    \item \textbf{\spiralagent}: Advanced agent employing Monte Carlo Tree Search (MCTS) assisted by an LLM-based Planner and Simulator to generate observations. The World-Model enables predictive simulation of tool outputs before actual execution.
    
\end{enumerate}

\subsection{World Model Configuration}
\label{modelselection}

We conducted a systematic evaluation across multiple agent and world model combinations. Specifically, we selected \texttt{gpt-oss-120b}, \texttt{claude-sonnet-4.6} as LLM based world models. And evaluated \texttt{Arctic-AWM-4B} as agent world model trained on MCP environment as introduced in \cite{wang2026agent}. This is the first MCP based world model series we came across in literature. Due to size and deployment complications and unavailability in popular LLM deployment providers, we stick to the smaller version of this series in our experiments. We opted \texttt{gpt-oss-120b} as planner baseline from our model set. Given the extensive human study done in MCP-Bench\cite{wang2026mcpbench}, we used \texttt{o4\_mini} as judge for evaluation. Overall, the setup resulted in 7 distinct configurations enabling comprehensive analysis of interactions among world model capabilities.

\begin{table*}[t]
\centering
\caption{Evaluation results comparing world-model-augmented agents against the ReAct baseline.
         Scores are reported as percentages~(\%; higher is better).
         Group scores are the mean of their two sub-dimensions;
         Overall $= (\text{Task Completion} + \text{Tool Selection} + \text{Planning Eff.}) / 3$.
         $\dag$~denotes the no-world-model baseline.
         \textbf{Bold}: best per column; \underline{underline}: second best;
         \colorbox{wingray}{shading}: improvement over baseline on that specific sub-dimension.}
\label{tab:wm-benchmark}
\setlength{\tabcolsep}{5pt}
\renewcommand{\arraystretch}{1.3}
\resizebox{\textwidth}{!}{%
\begin{tabular}{llcccccccccc}
\toprule
\multirow{3}{*}{\textbf{Agent}} &
\multirow{3}{*}{\textbf{World Model}} &
\multicolumn{2}{c}{\textbf{Task Completion}} &
\multicolumn{2}{c}{\textbf{Tool Selection}} &
\multicolumn{2}{c}{\textbf{Planning Effectiveness}} &
\multicolumn{4}{c}{\textbf{Aggregate}} \\
\cmidrule(lr){3-4}\cmidrule(lr){5-6}\cmidrule(lr){7-8}\cmidrule(lr){9-12}
& &
\makecell{Task\\Fulfillment} & \makecell{Ground-\\ing} &
\makecell{Tool\\Approp.} & \makecell{Param.\\Accuracy} &
\makecell{Dep.\\Awareness} & \makecell{Parallel.\\Effic.} &
\makecell{Task\\Compl.} & \makecell{Tool\\Sel.} & \makecell{Planning\\Effec.} &
\textbf{Overall} \\
\midrule
 
%--- Baseline ---
ReAct$^\dag$        & ---
  & \best{46.8} & 36.6
  & 41.5 & 31.3
  & \best{40.5} & 19.8
  & \best{41.7} & 36.4 & 30.1 & 36.1 \\
\midrule
 
%--- ReAct Plan variants ---
\reactplanagent  & \textsc{gpt-oss-120b}
  & 38.5               & \win{38.7}
  & \win{42.5}         & \win{\second{47.8}}
  & 36.9               & \win{20.6}
  & 38.6               & \win{45.2}  & 28.8             & \win{37.5} \\
 
\reactplanagent  & \textsc{claude-sonnet}
  & 31.4               & \win{\second{47.2}}
  & \win{\second{53.1}}& \win{\best{65.9}}
  & 27.1               & \win{\second{29.9}}
  & 39.3               & \win{\second{59.5}}  & 28.5    & \win{42.4} \\
 
\reactplanagent  & \textsc{awm-4B}
  & 33.7               & \win{42.6}
  & \win{45.2}         & \win{40.4}
  & 29.6               & \win{26.8}
  & 38.2               & \win{42.8}  & 28.2             & \win{36.4} \\
\midrule
 
%--- SPIRAL variants ---
\spiralagent      & \textsc{gpt-oss-120b}
  & 32.8               & \win{\best{51.0}}
  & \win{\best{59.6}}  & \win{61.0}
  & \second{33.6}      & \win{\best{30.9}}
  & \win{\second{41.9}}& \win{\best{60.3}}  & \win{\best{32.3}} & \win{\best{44.8}} \\
 
\spiralagent      & \textsc{claude-sonnet}
  & \second{34.5}      & \win{47.5}
  & \win{55.3}         & \win{\second{58.4}}
  & 33.3               & \win{27.5}
  & \win{41.0}         & \win{56.9}  & \win{\second{30.4}}   & \win{\second{42.8}} \\
 
\spiralagent      & \textsc{awm-4B}
  & 31.6               & 26.6
  & 33.6               & \win{48.4}
  & 20.5               & \win{20.2}
  & 29.1               & \win{41.0}  & 20.3             & 30.1 \\
 
\bottomrule
\end{tabular}%
}
 
\smallskip
\raggedright
\footnotesize
\textit{Notes:}
Task Completion $=$ (Task Fulfillment $+$ Grounding) / 2.
Tool Selection $=$ (Tool Appropriateness $+$ Parameter Accuracy) / 2.
Planning Effectiveness $=$ (Dependency Awareness $+$ Parallelism \& Efficiency) / 2.
Note that Planning Effectiveness averages Dependency Awareness and Parallelism \& Efficiency, which exhibit opposing trends across agents (see text).
SPIRAL: Structured Planning with Iterative Reflection and Lookahead.
AWM: Agent World Model (4B fine-tune).
LLM world models use \textsc{gpt-oss-120b} (same backbone as planner) or \textsc{claude-sonnet-4.6} (external).
\end{table*}

\subsection{Evaluation Metrics and Results}

To evaluate the influence of World Models in selecting the correct set of tools for each sub-task, specifically by capturing tool behavior and their underlying dependencies, we utilize a hierarchical evaluation framework from the MCP-Bench framework:

\begin{itemize}
    \item \textbf{Task Completion:} Aggregation of task's requirement fulfillment measure and groundness of final answer claims in tool outputs.
    \item \textbf{Task Selection:} Aggregation of metrics measuring appropriate tool choice for subtasks and accuracy and complete parameter selection for tool calls.
    \item \textbf{Planning Effectiveness:} Aggregation of metrics measuring parallelism in tool execution and ordered execution of tool dependency chains.
    \item \textbf{Overall Score (High-level):} This serves as the primary measure of task success rate. It is synthesized from three sub-metrics\cite{wang2026mcpbench}: \textit{Task Completion}, \textit{Tool Selection}, and \textit{Planning Effectiveness}.
\end{itemize} 

The experimental results, summarized in Table~\ref{tab:wm-benchmark}, evaluate the efficacy of World Model infusion against the \reactagent\ baseline with \texttt{gpt-oss-120b} planner model. A primary finding is that while \reactagent\ benefits from full-environment access and inherent error-recovery loops, both \reactplanagent\ and \spiralagent\ achieve better overall performance using almost all world model configurations.

\spiralagent\ + \texttt{gpt-oss-120b}-WM emerges as the clear winner at 44.8\%, followed by \reactplanagent\ + \texttt{claude-sonnet-4.6}-WM at 42.4\%. \texttt{AWM-4B}-WM, a purpose-built MCP world model is not as effective as general-purpose LLMs across most configurations. However, all of the WM infused agents bettered the baseline \reactagent\ agent in Parameter Accuracy, Parallel Efficiency and Tool Selection. This trend is consistent across multiple experiment runs. On the other hand, \reactagent\ remained strongest in Task Fulfillment, Dependency Awareness and Task Completion. This indicates that world models are helping to improve tool selection and parameter selection but not effective in completing the task. However, this observation reveals a measurement gap that the current evaluation is gearing towards task completeness rather than real intentions in world model infusion such as reducing failures in tool call executions, avoiding unnecessary tool calls, myopic agent behaviors etc. More insights on gaps in evaluation are discussed in Section~\ref{sec:eval}.

The baseline \reactagent\ agent consumed 49K tokens on average per task, the lowest versus rest of the configurations. \spiralagent\ +\texttt{Arctic-AWM-4B-WM}-WM consumed the highest number of tokens at ~302K on average per task representing increase of 5x over the baseline. Total token consumption across all the tasks range from 745K to 7M, highlighting the computation cost trade-offs of different architectural approaches. Although \texttt{Arctic-AWM-4B-WM} is small, it recorded ~3x tokens with both \spiralagent\ and \reactplanagent\ . Detailed cost analysis is done in Appendix~\ref{appendix:addanalysis}.

\section{Evaluation Gap Analysis and New Metrics}
\label{sec:eval}

Task fulfillment and dependency awareness are inherently dynamic and therefore harder to improve through simulator-driven proactive planning alone. Two metrics, tool call success rate, a metric measuring success or failure of tool calls, and number of tool calls in a trajectory are not used in Table\ref{tab:wm-benchmark}. These metrics provide important additional evidence about execution quality. \reactagent\ has the highest Task Fulfillment (46.8) yet the lowest tool call success rate (77.7). This suggests that it often reaches the correct outcome despite making more failed or inefficient tool calls along the way. 

To better capture this trade-off, we define a new metric, Execution Quality as the average of tool call success rate and the normalized average number of tool calls. We compute the normalized average number of tool calls relative to all experimental runs as \((\mathrm{max\_avg\_calls} - \mathrm{agent\_avg\_calls}) \,/\, 
(\mathrm{max\_avg\_calls} - \mathrm{min\_avg\_calls})\) and is expected to range from 0-100. This metric penalizes excessive retries and rewards efficient execution, making it more reflective of practical deployment settings where unnecessary tool calls increase latency and cost. Low the average tool calls, high tool call success rate and task completion is ideal outcome. Finally, we compute adjusted overall score accounting Execution Quality score.

\begin{table*}[t]
\centering
\caption{Evaluation results comparing world-model-augmented agents against the ReAct baseline.
         Scores are reported as percentages~(\%; higher is better).
         Group scores are the mean of their two sub-dimensions.
         Execution Quality $= (\text{Tool Call Success} + \widehat{\text{Avg Tool Calls}}) / 2$,
         where $\widehat{\text{Avg Tool Calls}}$ is min-max normalised and inverted (fewer calls $=$ higher score).
         Overall $= (\text{Task Completion} + \text{Tool Selection} + \text{Planning Eff.} + \text{Execution Quality}) / 4$.
         $\dag$~denotes the no-world-model baseline.
         \textbf{Bold}: best per column; \underline{underline}: second best;
         \colorbox{wingray}{shading}: improvement over baseline on that specific sub-dimension.}
\label{tab:wm-new-metric}
\setlength{\tabcolsep}{4pt}
\renewcommand{\arraystretch}{1.3}
\resizebox{\textwidth}{!}{%
\begin{tabular}{llccccccccccccccc}
\toprule
\multirow{3}{*}{\textbf{Agent}} &
\multirow{3}{*}{\textbf{World Model}} &
\multicolumn{2}{c}{\textbf{Task Completion}} &
\multicolumn{2}{c}{\textbf{Tool Selection}} &
\multicolumn{2}{c}{\textbf{Planning Effectiveness}} &
\multicolumn{2}{c}{\textbf{Execution Quality}} &
\multicolumn{5}{c}{\textbf{Aggregate}} \\
\cmidrule(lr){3-4}\cmidrule(lr){5-6}\cmidrule(lr){7-8}\cmidrule(lr){9-10}\cmidrule(lr){11-15}
& &
\makecell{Task\\Fulfil.} & \makecell{Ground-\\ing} &
\makecell{Tool\\Approp.} & \makecell{Param.\\Accuracy} &
\makecell{Dep.\\Aware.} & \makecell{Parallel.\\Effic.} &
\makecell{Tool Call\\Success} & \makecell{Avg Tool\\Calls$\downarrow$} &
\makecell{Task\\Compl.} & \makecell{Tool\\Sel.} & \makecell{Planning\\Effec.} & \makecell{Exec.\\Quality} &
\textbf{Overall} \\
\midrule
 
%--- Baseline ---
ReAct$^\dag$ & ---
  & \textbf{46.8} & 36.6
  & 41.5 & 31.2
  & \textbf{40.5} & 19.8
  & 77.7 & 0.0
  & 41.7 & 36.4 & 30.1 & 38.9 & 36.8 \\
\midrule
 
%--- WM agents, sorted by new Overall desc ---
\spiralagent & \textsc{gpt-oss-120b}
  & 32.8 & \win{\textbf{51.0}}
  & \win{\textbf{59.6}} & \win{\underline{61.0}}
  & \underline{33.6} & \win{\textbf{30.9}}
  & \win{\textbf{100.0}} & \win{82.8}
  & \win{\textbf{41.9}} & \win{\textbf{60.3}} & \win{\textbf{32.2}} & \win{\underline{91.4}} & \win{\textbf{56.5}} \\
 
\spiralagent\ & \textsc{claude-sonnet}
  & \underline{34.5} & \win{\underline{47.5}}
  & \win{\underline{55.3}} & \win{58.4}
  & 33.3 & \win{27.5}
  & \win{\textbf{100.0}} & \win{79.9}
  & \underline{41.0} & \win{56.9} & \win{\underline{30.4}} & \win{89.9} & \win{\underline{54.6}} \\
 
\reactplanagent\ & \textsc{claude-sonnet}
  & 31.4 & \win{47.2}
  & \win{53.1} & \win{\textbf{65.9}}
  & 27.1 & \win{\underline{29.9}}
  & \win{\textbf{100.0}} & \win{78.4}
  & 39.3 & \win{\underline{59.5}} & 28.5 & \win{89.2} & \win{54.1} \\
 
\reactplanagent\ & \textsc{awm-4B}
  & 33.7 & \win{42.6}
  & \win{45.2} & \win{40.4}
  & 29.6 & \win{26.8}
  & \win{\textbf{100.0}} & \win{\underline{85.8}}
  & 38.2 & \win{42.8} & 28.2 & \win{\textbf{92.9}} & \win{50.5} \\
 
\reactplanagent\ & \textsc{gpt-oss-120b}
  & \textbf{38.5} & \win{38.7}
  & \win{42.5} & \win{47.8}
  & \textbf{36.9} & \win{20.6}
  & \win{\textbf{100.0}} & \win{75.4}
  & 38.6 & \win{45.2} & 28.8 & \win{87.7} & \win{50.0} \\
 
\spiralagent\ & \textsc{awm-4B}
  & 31.6 & 26.6
  & 33.6 & \win{48.4}
  & 20.5 & \win{20.2}
  & \win{\textbf{100.0}} & \win{\textbf{100.0}}
  & 29.1 & \win{41.0} & 20.3 & \win{\textbf{100.0}} & \win{47.6} \\
 
\bottomrule
\end{tabular}%
}
 
\smallskip
\raggedright
\footnotesize
\textit{Notes:}
Task Completion $=$ (Task Fulfillment $+$ Grounding) / 2.
Tool Selection $=$ (Tool Appropriateness $+$ Parameter Accuracy) / 2.
Planning Effectiveness $=$ (Dependency Awareness $+$ Parallelism \& Efficiency) / 2.
Execution Quality $=$ (Tool Call Success $+$ $\widehat{\text{Avg Tool Calls}}$) / 2,
where $\widehat{\text{Avg Tool Calls}}$ is min-max normalised over all agents and inverted so that fewer calls score higher.
Overall $=$ (Task Completion $+$ Tool Selection $+$ Planning Effectiveness $+$ Execution Quality) / 4.
SPIRAL+AWM's perfect Execution Quality score reflects its low step count and should be interpreted alongside its Task Completion and Planning scores (see text).
SPIRAL: Structured Planning with Iterative Reflection and Lookahead.
AWM: Agent World Model (4B fine-tune).
LLM world models use \textsc{gpt-oss-120b} (same backbone as planner) or \textsc{claude-sonnet-4.6} (external).
\end{table*}

From experimental results in Table \ref{tab:wm-new-metric}, \reactagent\ is observed to have the highest average number of tool calls (7.04) and so the normalized score 0. This indicates that it frequently makes unsuccessful tool calls and compensates through repeated retries, while still eventually completing the task requirements. In contrast, this proposed Execution Quality metric better distinguishes agents that solve tasks efficiently from those that succeed only after incurring substantial execution overhead. Additional experimental analysis on computation etc. is provided in Appendix \ref{appendix:token_usage_overall_gpt}

\section{Ablation study}
\label{sec:ablation}

We conduct an ablation study to isolate the contribution of explicit world
models from planner capacity. Our research question is: \emph{can a stronger
planner compensate for the absence of an explicit world model?} To investigate
this, we replace the \textsc{gpt-oss-120b} planner with
\textsc{claude-sonnet-4.6}, a substantially more capable model with strong
in-context reasoning. The accuracy results in Table~\ref{tab:wm-ablation}
showed that \reactagent\ + \textsc{claude-sonnet-4.6} indeed improved over its
\textsc{gpt-oss-120b} counterpart, particularly in tool-call success rate.
The efficiency picture in Table~\ref{tab:efficiency_combined} tells a
different story.

\begin{table}[h!]
\centering
\small
\setlength{\tabcolsep}{4pt}
\caption{Efficiency metrics by planner. Rounds and tool calls are per-task averages;
execution time is in seconds. The \textsc{claude-sonnet-4.6} baseline issues
29.78 tool calls per task on average, against 1.12--7.91 for world-model-augmented
agents under the same planner.}
\label{tab:efficiency_combined}
\begin{tabular}{l rr rr rr}
\toprule
 & \multicolumn{2}{c}{\textbf{Rounds}} & \multicolumn{2}{c}{\textbf{Tool calls}} & \multicolumn{2}{c}{\textbf{Exec.\ time (s)}} \\
\cmidrule(lr){2-3} \cmidrule(lr){4-5} \cmidrule(lr){6-7}
\textbf{Method} & \textsc{gpt-oss} & \textsc{claude} & \textsc{gpt-oss} & \textsc{claude} & \textsc{gpt-oss} & \textsc{claude} \\
\midrule
\reactagent\ (baseline)                  & 7.42 & 14.13 &  7.04 & 29.78 &  63.7 & 214.9 \\
\midrule
\reactplanagent\ + gpt-oss-120b-WM       & 2.83 &  7.12 &  2.83 &  7.12 & 116.7 & 130.9 \\
\reactplanagent\ + claude-sonnet-4.6-WM  & 2.67 &  6.91 &  2.67 &  6.91 &  91.5 & 229.3 \\
\reactplanagent\ + Arctic-AWM-4B-WM      & 2.25 &  7.91 &  2.25 &  7.91 & 277.6 & 272.6 \\
\spiralagent\ + gpt-oss-120b-WM          & 2.42 &  1.92 &  2.42 &  1.92 &  65.2 &  88.0 \\
\spiralagent\ + claude-sonnet-4.6-WM     & 2.62 &  1.83 &  2.58 &  1.83 &  79.1 & 114.0 \\
\spiralagent\ + Arctic-AWM-4B-WM         & 1.54 &  1.25 &  1.46 &  1.12 & 570.9 & 720.0 \\
\bottomrule
\end{tabular}
\end{table}

\paragraph{The stronger planner trades efficiency for accuracy.}
The \textsc{claude-sonnet-4.6} baseline issues 29.78 tool calls per task on
average, more than 4$\times$ the \textsc{gpt-oss-120b} baseline (7.04) and
6--26$\times$ the world-model-augmented agents under either planner (1.12--7.91).
The accompanying execution time grows from 63.7s to 214.9s, a 3.4$\times$
slowdown, despite the underlying tasks being identical. The stronger planner is
not simply faster or more accurate per call; it is exploring a much wider
action space.

\paragraph{The planner's batching strategy also changes qualitatively.}
Beyond magnitude, the rounds-to-tool-calls ratio flips between the two baselines.
Under \textsc{gpt-oss-120b}, \reactagent\ issues roughly one tool call per round
(7.42 rounds, 7.04 calls). Under \textsc{claude-sonnet-4.6}, the same agent
issues 2.1 tool calls per round on average (14.13 rounds, 29.78 calls),
indicating that the stronger planner not only retries more but speculatively
batches multiple parallel calls per reasoning step. This is a behavioral
difference, not just a tuning difference, and it is the mechanism behind the
inflated tool-call counts and execution times of the powerful baseline.

\paragraph{World models constrain the powerful planner.}
When \textsc{claude-sonnet-4.6} is paired with a world model, the aggressive
exploration is suppressed: SPIRAL variants drop to 1.83--1.92 tool calls and
\reactplanagent\ variants to 6.91--7.91, comparable to or below the world-model
configurations under the weaker planner. The world model's simulation step
forces the planner to commit to a vetted plan rather than probe the live
environment, converting speculative breadth into focused execution. This is the
mechanism by which Execution Quality improves: not because the planner becomes
weaker, but because the world model constrains where it spends its budget.

\begin{table*}[t]
\centering
\caption{Evaluation results for \textsc{claude-sonnet-4.6} as the planner, comparing
         world-model-augmented agents against the ReAct baseline.
         Scores are reported as percentages~(\%; higher is better).
         Group scores are the mean of their two sub-dimensions.
         Overall$_\text{orig}$ $= (\text{Task Compl.} + \text{Tool Sel.} + \text{Planning Eff.}) / 3$.
         Execution Quality $= (\text{Tool Call Success} + \widehat{\text{Avg Tool Calls}}) / 2$,
         where $\widehat{\text{Avg Tool Calls}}$ is min-max normalised and inverted (fewer calls $=$ higher score).
         Overall$_\text{new}$ $= (\text{Task Compl.} + \text{Tool Sel.} + \text{Planning Eff.} + \text{Exec. Qual.}) / 4$.
         $\dag$~denotes the no-world-model baseline.
         \textbf{Bold}: best per column; \underline{underline}: second best;
         \colorbox{wingray}{shading}: improvement over baseline on that specific sub-dimension.}
\label{tab:wm-ablation}
\setlength{\tabcolsep}{4pt}
\renewcommand{\arraystretch}{1.3}
\resizebox{\textwidth}{!}{%
\begin{tabular}{ll cc cc cc cccc cc c c}
\toprule
\multirow{3}{*}{\textbf{Agent}} &
\multirow{3}{*}{\textbf{World Model}} &
\multicolumn{2}{c}{\textbf{Task Completion}} &
\multicolumn{2}{c}{\textbf{Tool Selection}} &
\multicolumn{2}{c}{\textbf{Planning Effectiveness}} &
\multicolumn{4}{c}{\textbf{Aggregate}} &
\multicolumn{2}{c}{\textbf{Execution Quality}} &
\multicolumn{1}{c}{\textbf{Exec.}} &
\multicolumn{1}{c}{\textbf{New}} \\
\cmidrule(lr){3-4}\cmidrule(lr){5-6}\cmidrule(lr){7-8}\cmidrule(lr){9-12}\cmidrule(lr){13-14}
& &
\makecell{Task\\Fulfil.} & \makecell{Ground-\\ing} &
\makecell{Tool\\Approp.} & \makecell{Param.\\Accur.} &
\makecell{Dep.\\Aware.} & \makecell{Parallel.\\Effic.} &
\makecell{Task\\Compl.} & \makecell{Tool\\Sel.} & \makecell{Planning\\Effec.} &
\makecell{Overall\\$_\text{orig}$} &
\makecell{Tool Call\\Success} & \makecell{Avg Tool\\Calls$\downarrow$} &
\makecell{Qual.\\(\%)} &
\makecell{Overall\\$_\text{new}$} \\
\midrule
 
%--- Baseline ---
ReAct$^\dag$ & ---
  & \textbf{66.6} & \textbf{66.9}
  & 64.0 & 52.4
  & \textbf{59.2} & 30.4
  & \textbf{66.7} & 58.2 & \textbf{44.8} & \textbf{56.6}
  & 83.7 & 0.0 & 41.9 & 52.9 \\
\midrule
 
%--- WM agents sorted by new Overall desc ---
\reactplanagent\ & \textsc{gpt-oss-120b}
  & 41.8 & \textbf{58.4}
  & \win{\textbf{70.9}} & \win{\underline{59.8}}
  & \underline{50.2} & \win{\textbf{31.9}}
  & \underline{50.1} & \win{\textbf{65.3}} & \textbf{41.1} & \textbf{52.1}
  & \win{\textbf{100.0}} & \win{79.1} & \win{89.5} & \win{\textbf{61.5}} \\
 
\reactplanagent\ & \textsc{claude-sonnet}
  & \underline{48.2} & 52.7
  & \win{\underline{65.1}} & \win{58.9}
  & \textbf{50.3} & 26.5
  & \textbf{50.4} & \win{\underline{62.0}} & \underline{38.4} & \underline{50.3}
  & \win{\textbf{100.0}} & \win{79.8} & \win{89.9} & \win{\underline{60.2}} \\
 
\reactplanagent & \textsc{awm-4B}
  & 41.6 & \underline{56.0}
  & 60.5 & \win{\textbf{62.4}}
  & 46.0 & \underline{29.3}
  & 48.8 & \win{61.5} & 37.6 & 49.3
  & \win{\textbf{100.0}} & \win{76.3} & \win{88.2} & \win{59.0} \\
 
\spiralagent & \textsc{gpt-oss-120b}
  & 46.1 & 32.9
  & 38.2 & 48.6
  & 41.5 & 25.0
  & 39.5 & 43.4 & 33.2 & 38.7
  & \win{\textbf{100.0}} & \win{97.2} & \win{98.6} & \win{53.7} \\
 
\spiralagent & \textsc{claude-sonnet}
  & \textbf{51.0} & 32.9
  & 35.1 & 48.4
  & 38.2 & 22.9
  & 42.0 & 41.7 & 30.6 & 38.1
  & \win{\textbf{100.0}} & \win{\underline{97.5}} & \win{\underline{98.8}} & \win{53.3} \\
 
\spiralagent & \textsc{awm-4B}
  & 47.1 & 32.7
  & 29.7 & 43.4
  & 41.4 & 21.7
  & 39.9 & 36.5 & 31.5 & 36.0
  & \win{\textbf{100.0}} & \win{\textbf{100.0}} & \win{\textbf{100.0}} & 52.0 \\
 
\bottomrule
\end{tabular}%
}
 
\smallskip
\raggedright
\footnotesize
\textit{Notes:}
Task Completion $=$ (Task Fulfillment $+$ Grounding) / 2.
Tool Selection $=$ (Tool Appropriateness $+$ Parameter Accuracy) / 2.
Planning Effectiveness $=$ (Dependency Awareness $+$ Parallelism \& Efficiency) / 2.
Execution Quality $=$ (Tool Call Success $+$ $\widehat{\text{Avg Tool Calls}}$) / 2,
where $\widehat{\text{Avg Tool Calls}}$ is min-max normalised over all agents and inverted so that fewer calls score higher.
SPIRAL+AWM's perfect Execution Quality reflects its low step count; interpret alongside its Task Completion and Planning scores (see text).
SPIRAL: Structured Planning with Iterative Reflection and Lookahead.
AWM: Agent World Model (4B fine-tune).
LLM world models: \textsc{gpt-oss-120b} or \textsc{claude-sonnet-4.6}.
\end{table*}

\paragraph{Implication.}
The combination of powerful planners with low-quality simulations from a weaker world model proves undesirable. A stronger planner does not substitute for an explicit world model on the efficiency axis; if anything, the stronger the planner, the more an explicit world model matters. This is because the latency and cost of unconstrained exploration scale with both planner capability and the quality of simulations. We did not observe a conclusive ordering between homogeneous (planner = world-model backbone) and heterogeneous configurations in either Table~\ref{tab:wm-new-metric} or Table~\ref{tab:wm-ablation}; we leave this question to future work. Detailed token-usage and per-planner efficiency breakdowns are reported in Appendix~\ref{appendix:addanalysis}.

\section{Limitations}

While \textbf{MCP-Cosmos} demonstrates the efficacy of World Model infusion, several limitations remain. First, our evaluation is grounded in the \textbf{MCP-Bench} framework, which, while diverse, relies on a static snapshot of server environments. Real-world ecosystems are inherently dynamic, and our current World Models do not yet incorporate real-time online learning to adapt to evolving tool schemas or API behaviors. Second, the computational overhead of utilizing high-parameter models like \textsc{claude-sonnet-4.6} as the World Model is significant. While these models provide superior grounding, the cost-performance trade-off may be prohibitive for latency-sensitive or resource-constrained applications. Third, our evaluation metrics focus primarily on task success and execution quality, but do not yet capture the quality of intermediate reasoning steps or the interpretability of world model representations. 
Additionally, the Execution Quality metric relies on min-max normalization within each experimental cohort, making scores non-portable across different experiments or papers. However, this is by design rather than a limitation. When benchmarking the usefulness of world models in agent planning, the evaluation must be conducted against a specific planner baseline. So, the proposed metric is suitable for world model evaluation rather than general purpose evaluation.

\section{Related Work}

\subsection{World Models}

World models learn predictive representations of environment dynamics that enable planning, reasoning, and decision-making. They have been studied across reinforcement learning\cite{wu2025rlvrworld}, robotics, and multimodal generation\cite{zhao2025from}\cite{brito2025world}\cite{yang2025mindjourney}, with recent work focusing on scalable architectures, test-time rollout, and systematic evaluation. Wang et al.~\cite{wang2025sampo} explore autoregressive and spatial world models for video and robotic domains, emphasizing scalable spatiotemporal prediction. Chen et al.~\cite{chen2025learning} study world model construction for interactive video generation, addressing how environment dynamics can be learned directly from video trajectories. Yang et al.~ leverage world models to roll out imaginary trajectories at test time, enabling spatial reasoning and search-based inference using separate vision–language models for exploration and answer generation.

World models research in MCP environments gaining attention recently. Wang et al.~\cite{wang2026agent} introduced automated tool-use environment generation pipeline and contributed 1000 ready-to-use environments. Further, the authors have contributed three RL trained models as well. As MCP is widely popular, exponentially growing agentic systems and introduction of new MCP based world models targetting production systems, there is need for evolution of world model specific evaluation methods in MCP environments.

\section{Conclusion}

We introduced \textbf{MCP-Cosmos}, an evaluation suite curated from the ecosystem-scale \textbf{MCP-Bench} to investigate the impact of world model infusion on agentic planning. While iterative frameworks like \reactagent\ have traditionally served as the performance ceiling due to their trial-and-error recovery, \textbf{\reactplanagent} and \textbf{\spiralagent}, when augmented with powerful world models, achieve better tool selection and parameter accuracy, and increase the possibility of parallel execution due to their proactive planning strategy. The proposed Execution Quality metric improves the evaluation of world models by quantifying their ability to guide efficient, targeted tool usage with minimal exploratory overhead. While the \texttt{Arctic-AWM-4B}-WM trained on 1000 MCP environments \cite{wang2026agent} is not as effective as general-purpose LLMs, the proposed evaluation metrics and the MCP-Cosmos framework can aid in evaluating these upcoming MCP world models. Future work could explore the impact of world models on write operations, destructive tools, environment state management, simulation fidelity analysis, and domain-specific adaptation strategies.

\bibliographystyle{plainnat}
\bibliography{reference}

\clearpage
\appendix

\section{Technical appendices and supplementary material}
\label{appendix:addanalysis}

\subsection{System Requirements}
The experiments are run on 2021 M1 Max Macbook Pro with 32GB memory. Although most of the models are deployed in cloud, Arctic AWM 4B is deployed natively on the above Mac. Although the code is general enough to run on Mac, Windows or Linux, it is tested primarily on M1 Mac.

\subsection{Token Usage Analysis - gpt-oss-120b Planner}
\label{appendix:token_usage_overall_gpt}

This appendix presents comprehensive token usage statistics across all evaluated agent configurations using the \texttt{gpt-oss-120b} planner. The data includes both average per-task metrics and cumulative totals across all benchmark tasks. Results are reported using pass@k=4.

\begin{table}[htbp]
\centering
\small
\setlength{\tabcolsep}{4pt}
\caption{Overall token usage by agent configuration with the \texttt{gpt-oss-120b} planner. Average values are per task; total values are aggregated over the full benchmark.}
\label{tab:token_usage_overall}
\begin{tabular}{l rrr rrr}
\toprule
 & \multicolumn{3}{c}{\textbf{Average (per task)}} & \multicolumn{3}{c}{\textbf{Total (benchmark)}} \\
\cmidrule(lr){2-4} \cmidrule(lr){5-7}
\textbf{Method} & \textbf{Output} & \textbf{Prompt} & \textbf{Total} & \textbf{Output} & \textbf{Prompt} & \textbf{Total} \\
\midrule
\reactagent                              &  4{,}972 &  45{,}024 &  \textbf{49{,}995} & 119{,}320 & 1{,}080{,}568 & 1{,}199{,}888 \\
\reactplanagent\ + claude-sonnet-4.6-WM  &  7{,}801 &  75{,}023 &  82{,}825 &  70{,}210 &    675{,}211 &    745{,}421 \\
\spiralagent\ + claude-sonnet-4.6-WM     &  7{,}469 &  81{,}507 &  88{,}976 & 179{,}266 & 1{,}956{,}162 & 2{,}135{,}428 \\
\spiralagent\ + gpt-oss-120b-WM          &  9{,}002 &  81{,}435 &  90{,}437 & 216{,}039 & 1{,}954{,}442 & 2{,}170{,}481 \\
\reactplanagent\ + Arctic-AWM-4B-WM      & 12{,}704 & 102{,}082 & 114{,}786 & 304{,}895 & 2{,}449{,}958 & 2{,}754{,}853 \\
\reactplanagent\ + gpt-oss-120b-WM       & 13{,}679 & 101{,}583 & 115{,}262 & 109{,}430 &    812{,}665 &    922{,}095 \\
\spiralagent\ + Arctic-AWM-4B-WM         & 33{,}756 & 268{,}152 & 301{,}908 & 810{,}149 & 6{,}435{,}641 & 7{,}245{,}790 \\
\bottomrule
\end{tabular}
\end{table}

\paragraph{Token usage analysis.}
The baseline \reactagent\ is the most token-efficient configuration with an average of 49{,}995 tokens per task. World-model-augmented agents incur 66\%--504\% higher per-task token consumption, with \spiralagent\ + Arctic-AWM-4B-WM at the upper end (301{,}908 tokens, $\approx$6$\times$ the baseline). Across the benchmark, total token consumption ranges from 745K (\reactplanagent\ + claude-sonnet-4.6-WM) to 7.2M (\spiralagent\ + Arctic-AWM-4B-WM), highlighting the computational trade-off introduced by world-model simulation. Prompt tokens dominate output tokens by roughly 9$\times$ across all methods, indicating that token cost is driven primarily by context retention and prompt construction rather than generated responses.

The token data corroborates the efficiency findings reported in the main body (Section~\ref{sec:ablation}, Table~\ref{tab:efficiency_combined}): world models reduce reasoning iterations but introduce simulation overhead, and the choice between LLM-based and AWM-based world models reflects a trade-off between capability and inference cost.

\subsection{Token Usage Analysis - claude-sonnet-4.6 Planner}
\label{appendix:token_usage_analysis_claude}

This appendix presents token usage statistics across all evaluated agent configurations using the \texttt{claude-sonnet-4.6} planner.

\begin{table*}[t]
\centering
\small
\setlength{\tabcolsep}{4pt}
\caption{Overall token usage across benchmark configurations with the \texttt{claude-sonnet-4.6} planner. Average values are per task; total values are aggregated over the full benchmark.}
\label{tab:token_usage_overall_claude}
\begin{tabular}{l rrr rrr}
\toprule
 & \multicolumn{3}{c}{\textbf{Average (per task)}} & \multicolumn{3}{c}{\textbf{Total (benchmark)}} \\
\cmidrule(lr){2-4} \cmidrule(lr){5-7}
\textbf{Method} & \textbf{Output} & \textbf{Prompt} & \textbf{Total} & \textbf{Output} & \textbf{Prompt} & \textbf{Total} \\
\midrule
\reactplanagent\ + Arctic-AWM-4B-WM      &  8{,}512 &  87{,}481 & \textbf{95{,}993} &  93{,}633 &    962{,}291 &  1{,}055{,}924 \\
\spiralagent\ + gpt-oss-120b-WM          &  5{,}069 &  93{,}698 &  98{,}767 & 121{,}666 &  2{,}248{,}745 &  2{,}370{,}411 \\
\reactagent                              &  5{,}238 &  93{,}888 &  99{,}126 & 120{,}467 &  2{,}159{,}430 &  2{,}279{,}897 \\
\spiralagent\ + claude-sonnet-4.6-WM     &  5{,}249 &  97{,}880 & 103{,}129 & 125{,}987 &  2{,}349{,}114 &  2{,}475{,}101 \\
\reactplanagent\ + gpt-oss-120b-WM       &  5{,}968 & 100{,}068 & 106{,}036 &  47{,}747 &    800{,}545 &    848{,}292 \\
\reactplanagent\ + claude-sonnet-4.6-WM  &  7{,}262 & 116{,}264 & 123{,}526 &  79{,}877 &  1{,}278{,}906 &  1{,}358{,}783 \\
\spiralagent\ + Arctic-AWM-4B-WM         & 28{,}254 & 658{,}519 & 686{,}772 & 678{,}092 & 15{,}804{,}446 & 16{,}482{,}538 \\
\bottomrule
\end{tabular}
\end{table*}

\paragraph{Token usage analysis.}
Under the \texttt{claude-sonnet-4.6} planner, \reactplanagent\ + Arctic-AWM-4B-WM is the most token-efficient configuration overall (95{,}993 average total tokens per task), slightly below the baseline \reactagent\ (99{,}126), suggesting that adding planning with AWM can improve token efficiency without increasing cost relative to a powerful baseline.

\spiralagent\ + gpt-oss-120b-WM, \spiralagent\ + claude-sonnet-4.6-WM, and \reactagent\ form a tight cluster in the 99K--103K range, indicating broadly similar inference cost profiles despite differing execution strategies. \reactplanagent\ variants with LLM world models show measurable overhead (106K--124K), reflecting the additional cost of explicit planning and world-model prompting.

As in the gpt-oss-120b setting, prompt tokens dominate output tokens (e.g., 93{,}888 vs 5{,}238 for \reactagent), confirming that token cost is driven by prompt construction rather than generation. \spiralagent\ + Arctic-AWM-4B-WM remains a clear outlier at 686{,}772 average tokens per task and 16.48M cumulatively---an order of magnitude above other methods---and should be characterized as a high-cost configuration regardless of its quality scores.

\section{Algorithm}
\label{appendix:algorithm}

As shown in Algorithm~\ref{alg:world_model_planning}, the planning phase (lines 6--15) uses the world model to simulate potential action sequences without incurring real-world execution cost. Starting from an initial state $s_0$ derived from the task instruction $u$, the agent iteratively generates actions $a_t$ using the planning policy $\pi_{\text{plan}}$ and the relevant list of MCP tools from MCP servers. For each action, the world model predicts a simulated (pseudo) observation $\tilde{o}_t$ that approximates the expected outcome without actual execution. This simulated observation need not be in the same format or capacity as the actual observation: it can be a summary explanation of the possible outcome, an example demonstrating the structure of an actual observation, simulated data resembling the patterns in actual observations, or information highlighting the consequences of an actual observation. These action and simulated-observation pairs are accumulated in the world-model trajectory $\tau_{\text{wm}}$, allowing the agent to explore multiple potential paths efficiently.

\begin{algorithm}[h!]
\caption{WM Infused Planning and Execution}
\label{alg:world_model_planning}
\begin{algorithmic}[1]
\Require Task instruction $u$, maximum steps $T_{\max}$
\Ensure Final Answer $\textit{answer}$, Execution Plan $\mathcal{P}$, execution trajectory $\tau$

\Function{MultiTurnWithWorldModel}{$u, T_{\max}, \text{WorldModel}$}
    \State $\mathcal{P} \gets \emptyset$ \Comment{Initialization}
    \State $\tau \gets \emptyset$
    \State $s_0 \gets \text{update}(u)$
    \State $\tau_{\text{wm}} \gets \emptyset$
    
    \For{$t = 0$ \textbf{to} $T_{\max}$}
        \State $a_t \gets \pi_{\text{plan}}(s_t)$
        \State $\tilde{o}_t \gets \text{WorldModel}(a_t)$ \Comment{Simulate pseudo observation}
        \State Update $\tau_{\text{wm}}$ with $\{(a_t, \tilde{o}_t)\}$
        \State $s_{t+1} \gets \text{update}(s_t, \tilde{o}_t)$
        \If{$\text{continue}_t = \texttt{false}$}
            \State \textbf{break}
        \EndIf
    \EndFor
    
    \State $\mathcal{P} \gets \text{select\_optimal\_plan}(\tau_{\text{wm}})$ \Comment{Select best plan from world-model explorations}
    
    \For{$a_t \in \mathcal{P}$}
        \State $o_t \gets \text{exec}(a_t)$
        \If{$o_t$ fails}
            \State $\mathcal{P} \gets \pi_{\text{plan\_adjust}}(\mathcal{P})$ \Comment{Optionally revise plan to mitigate failures}
            \State \textbf{continue}
        \EndIf
        \State Update $\tau$ with $\{(a_t, o_t)\}$
    \EndFor
    
    \State $\textit{answer} \gets \pi_{\text{final}}(\tau)$
    \State \Return $\textit{answer}, \mathcal{P}, \tau$
\EndFunction
\end{algorithmic}
\end{algorithm}

The state is updated based on simulated observations, enabling the agent to reason about future states and make informed decisions. This process continues until either the maximum number of steps $T_{\max}$ is reached or a termination condition is met (indicated by $\text{continue}_t = \texttt{false}$). The accumulated world-model trajectory $\tau_{\text{wm}}$ captures the agent's simulated exploration and serves as the basis for selecting an optimal execution plan via the \texttt{select\_optimal\_plan} function (line 15). This selection can use a non-deterministic policy model such as an LLM, or any deterministic selection algorithm such as reward-based MCTS~\cite{zhang2025spiralsymbolicllmplanning}. The selected final plan represents the sequence of tasks or tool calls to be executed along with their parameters.

Once a structured plan with tool calls is available, the execution phase (lines 16--23) carries out execution of each action in the actual environment. For each action $a_t$ in the plan, the agent executes and receives the real observation $o_t$. If an action fails during execution, the algorithm can optionally invoke a plan adjustment mechanism $\pi_{\text{plan\_adjust}}$ to modify the remaining plan, demonstrating robustness to execution failures. Due to the expensive nature of these revisions, we excluded the plan revision step from our benchmarking pipeline. At each step, successfully executed action--observation pairs are recorded in the execution trajectory $\tau$.

Upon completing execution of the plan, we generate the final answer using summarization or synthesis techniques. The algorithm returns three key outputs: the final answer, the execution plan $\mathcal{P}$, and the complete execution trajectory $\tau$, providing full transparency into both the planning and execution processes.

\section{MCP Task and Tool Distribution}
\label{sec:appendix_tasks}

This appendix provides the full listing of the 24 curated scenarios from the MCP-Bench dataset selected for the \textbf{MCP-Cosmos} evaluation suite. These tasks were chosen to emphasize ecosystem-scale complexity and multi-server coordination.

\subsection{High-Coordination Scenarios (3 Servers)}
The following 6 tasks require simultaneous grounding and dependency management across three distinct MCP servers:

\begin{itemize}
    \item \texttt{paper\_search\_call\_for\_papers\_wiki\_000}
    \item \texttt{paper\_search\_call\_for\_papers\_wiki\_002}
    \item \texttt{metropolitan\_mus\_huge\_icons\_wiki\_000}
    \item \texttt{metropolitan\_museum\_huge\_icons\_wiki\_001}
    \item \texttt{medical\_calculator\_wikipedia\_fruit\_000}
    \item \texttt{medical\_calculator\_wikipedia\_fruit\_001}
\end{itemize}

\subsection{Dual-Server Grounding Scenarios (2 Servers)}
The following 18 tasks evaluate bilateral server interactions and stability in handling fuzzy instructions:

\begin{itemize}
    \item \texttt{nixos\_context7\_000}
    \item \texttt{nixos\_context7\_001}
    \item \texttt{dex\_paprika\_okx\_exchange\_000}
    \item \texttt{dex\_paprika\_okx\_exchange\_001}
    \item \texttt{metropolitan\_museum\_wikipedia\_000}
    \item \texttt{metropolitan\_museum\_wikipedia\_001}
    \item \texttt{scientific\_computing\_math\_mcp\_000}
    \item \texttt{scientific\_computing\_math\_mcp\_001}
    \item \texttt{unit\_converter\_math\_mcp\_000}
    \item \texttt{unit\_converter\_math\_mcp\_001}
    \item \texttt{game\_trends\_reddit\_000}
    \item \texttt{game\_trends\_reddit\_001}
    \item \texttt{scientific\_computing\_unit\_converter\_000}
    \item \texttt{scientific\_computing\_unit\_converter\_001}
    \item \texttt{wikipedia\_paper\_search\_001}
    \item \texttt{wikipedia\_paper\_search\_003}
    \item \texttt{reddit\_dex\_paprika\_000}
    \item \texttt{reddit\_dex\_paprika\_001}
\end{itemize}

%%%%%%%%%%%%%%%%%%%%%%%%%%%%%%%%%%%%%%%%%%%%%%%%%%%%%%%%%%%%

\end{document}